\documentclass[letterpaper, 10 pt, conference]{ieeeconf}
\usepackage{PackageHeading}
\usepackage[absolute,overlay]{textpos}

\title{\Large
Identifying Optimal Launch Sites of High-Altitude Latex-Balloons using Bayesian Optimisation for the Task of Station-Keeping
\vspace{-4 mm}}

\author{Jack Saunders$^{1}$, Sajad Saeedi$^{2}$, Adam Hartshorne$^{1}$, Binbin Xu$^{3}$, Özgür \c{S}im\c{s}ek$^{1}$, Alan Hunter$^{1}$, and Wenbin Li$^{1}$
\vspace{-4 mm}
\thanks{This work is supported by the UKRI Centre for Doctoral Training in Accountable, Responsible \& Transparent AI (ART-AI), under UKRI grant number EP/S023437/1.}
\thanks{$^{1}$University of Bath, \{\protect\url{js3442, ath35, os435, A.J.Hunter, w.li}\}\protect\url{@bath.ac.uk}}
\thanks{
        $^{2}$Toronto Metropolitan University, \protect\url{s.saeedi@torontomu.ca}}
\thanks{
        $^{3}$University of Toronto, \protect\url{binbin.xu@utoronto.ca}}%
}

\begin{document}
\maketitle
\thispagestyle{empty}
\pagestyle{empty}

\begin{textblock*}{12cm}(4.75cm,1cm) 
  \centering
  \textcolor{red}{\small
  This work has been submitted to the IEEE International Conference on Intelligent Robots and Systems (IROS) for possible publication.  Copyright may be transferred without notice, after which this version may no longer be accessible.}
\end{textblock*}

\begin{abstract}
Station-keeping tasks for high-altitude balloons show promise in areas such as ecological surveys, atmospheric analysis, and communication relays.
However, identifying the optimal time and position to launch a latex high-altitude balloon is still a challenging and multifaceted problem.  For example, tasks such as forest fire tracking place geometric constraints on the launch location of the balloon.  Furthermore, identifying the most optimal location also heavily depends on atmospheric conditions.
We first illustrate how reinforcement learning-based controllers, frequently used for station-keeping tasks, can exploit the environment. 
This exploitation can degrade performance on unseen weather patterns and affect station-keeping performance when identifying an optimal launch configuration. Valuing all states equally in the region, the agent exploits the region's geometry by flying near the edge, leading to risky behaviours.  We propose a modification which compensates for this exploitation and finds this leads to, on average, higher steps within the target region on unseen data.
Then, we illustrate how Bayesian Optimisation (BO) can identify the optimal launch location to perform station-keeping tasks, maximising the expected undiscounted return from a given rollout.  We show BO can find this launch location in fewer steps compared to other optimisation methods.  Results indicate that, surprisingly, the most optimal location to launch from is not commonly within the target region.  Please find further information about our project at \href{https://sites.google.com/view/bo-lauch-balloon/}{https://sites.google.com/view/bo-lauch-balloon/}.
\end{abstract}

\section{Introduction}
Recently, high-altitude balloons have shown promise in applications such as environmental \cite{tironi_militarization_2021} and wildlife \cite{adams_coexisting_2020, wang_optical_2021} surveillance, communication relay \cite{bellemare_autonomous_2020, alsamhi_tethered_2019}, and atmospheric analysis \cite{alam_design_2018}. 
Furthermore, high-altitude balloons can sustain flights for many days \cite{sushko_low_2017, bellemare_autonomous_2020}.  Whereas in comparison, conventional unmanned aerial vehicles succumb to power and weight constraints, leading to reduced flight time \cite{saunders_autonomous_2024}.  Latex-balloon alternatives are currently being explored as a low-cost alternative to super-pressure balloons \cite{sushko_low_2017}.  Where super-pressure balloons depend on a high-strength plastic envelope to prevent expansion of the lifting gas, which can be cost-prohibitive for research purposes.
These passively actuated balloons are under-actuated, with direct control limited to the ascent rate.  Hence, their flight path is dictated by the direction of the wind and atmospheric conditions.

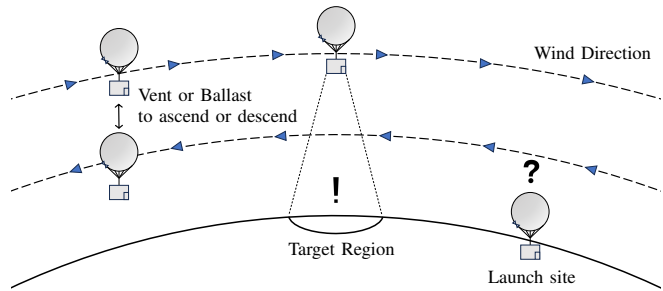
\begin{figure}
\centering
\vspace*{-3mm}
\begin{overpic}[width=\linewidth]{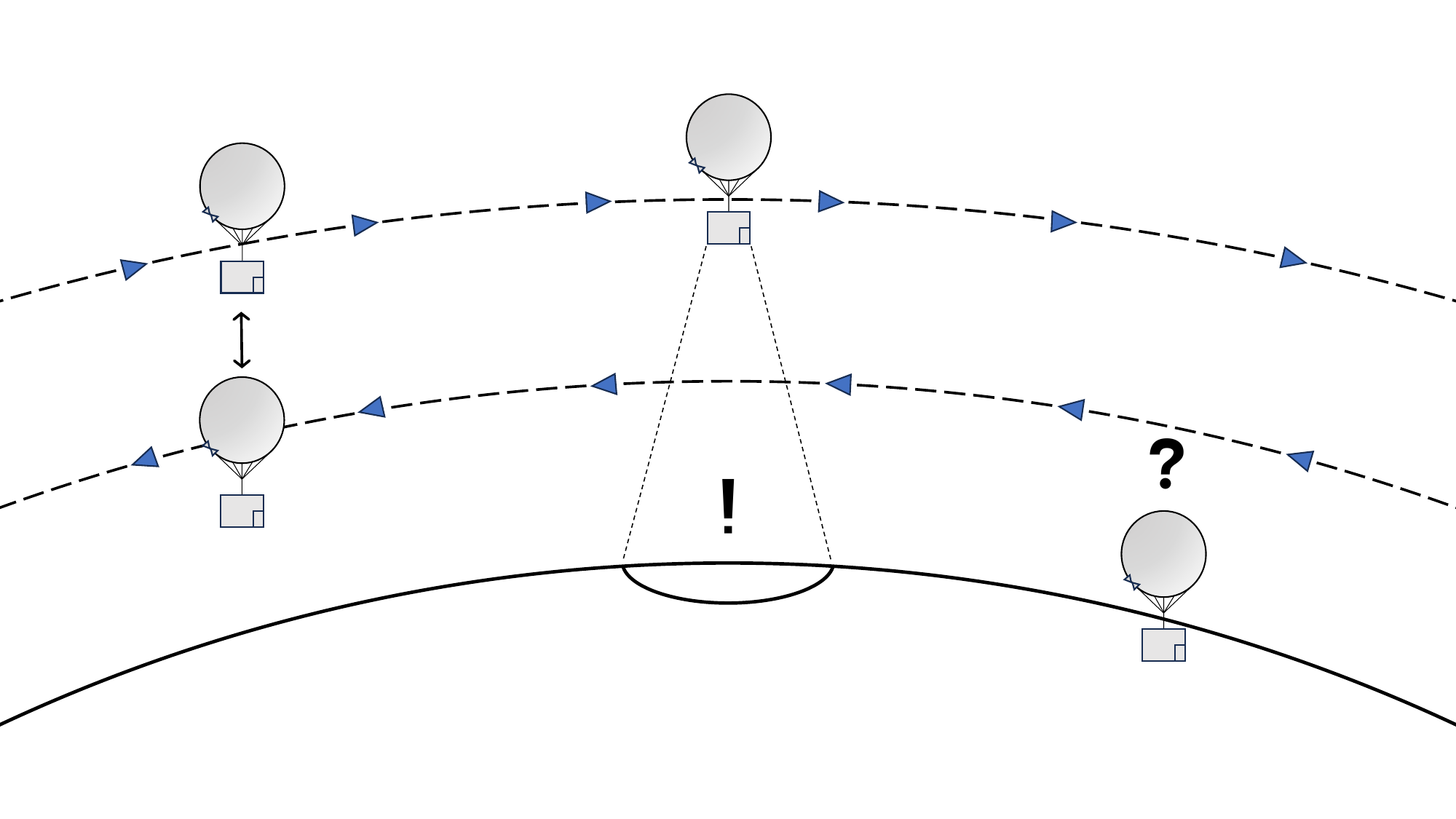}
\put(105, 85){\makebox(0,0){\begin{minipage}{4cm}
\scriptsize Vent or Ballast \\ to ascend or descend
\end{minipage}}}
\put(125, 30){\makebox(0,0){\scriptsize Target Region}}
\put(220, 105){\makebox(0,0){\scriptsize Wind Direction}}
\put(197, 20){\makebox(0,0){\scriptsize Launch site}}

\end{overpic}
\vspace*{-8mm}

\caption{{\small Here, we illustrate our objective to identify a launch configuration that maximises the duration within the target region.  Despite the challenges: of under-actuated navigation, the influence of wind and atmospheric conditions, and environment exploitation.}}
\label{fig:objective_intro}
\end{figure}

Reinforcement learning (RL) based controllers for station-keeping have increased in popularity, particularly due to the wind-forecast uncertainties and the non-linear relationship between actions and optimal states as a result of atmospheric conditions \cite{bellemare_autonomous_2020}.  This relationship leads to difficulties in applying search-based and model-based approaches for station-keeping tasks.  
However, RL-based controllers are known to lead to unintended behaviour that can emerge from under-defined reward functions \cite{amodei_concrete_2016}.  We show that agents trained for station-keeping can game the reward function by exploiting the geometries of the region, leading to poorer performance on out-of-distribution data. Additionally, this exploitation can degrade the performance of the identified optimal launch configuration.
Some previous reward functions \cite{saunders_resource-constrained_2023, bellemare_autonomous_2020} gave equal value to states within the region.  In some instances, this incentivised agents to fly along the boundary of the region, which allowed the agent to obtain more reward at the expense of risky behaviours.  Instead, we propose a modification which increases the reward closer to the target when inside the region using a Tanh function.  We find this modified reward function leads to, on average, higher steps within the 50\;km region on unseen data.

Researchers have previously established the temporal factors, such as diurnal cycles and seasonal variations, on station-keeping performance \cite{bellemare_autonomous_2020}.  Despite this, the effects of varying spatial positions on balloon performance have not been thoroughly investigated, and optimisation strategies accounting for spatial variability remain unexplored.  As a result, we illustrate how Bayesian Optimisation (BO) can locate launch configurations and identify latitude, longitude, and time to launch.  Furthermore, we also show that BO is capable of finding the launch configuration in almost half as many iterations compared to other optimisation approaches.  For our study, we focus on latex-based balloons, however the use of BO to find the launch configuration can be used for any balloon type.


Our contributions are as follows:
\underline{\textbf{(1)}} To our knowledge, we are the first to propose an optimization strategy for high-altitude balloon station-keeping.  Considering both spatial and temporal variability to determine the optimal launch configuration.
\underline{\textbf{(2)}} We propose a modified reward function that mitigates reward hacking for station-keeping tasks, and which increases maximum station-keeping performance.
\underline{\textbf{(3)}}~We then show that Bayesian Optimisation can identify the optimal launch location in fewer steps compared to other optimisation approaches.


\section{Literature Review}\label{sec:lit}
Researchers have proposed controllers to maintain a high-altitude balloon's position within a target region, also known as station-keeping.  These controllers exploit the varying wind directions at different altitudes to optimise the balloon's time spent in a designated area.

Du {\it et al.} introduces a geometric method to calculate the optimal altitude to reach in order to maximise the cosine similarity between the orientation of the balloon \cite{du_station-keeping_2019}.  The method, however, assumes stable wind fields, which is unpredictable.   Liu {\it et al.} extend this work by proposing their Time-Varying Range of Achievable Altitude Algorithm (TR3A).  This algorithm incorporates the bursting and over-pressure altitudes, which more accurately represent the range of altitudes the balloon can safely reach \cite{liu_increased_2022}.  The authors illustrated improved station-keeping performance by taking advantage of an increased range of altitudes.

Bellemare {\it et al.} argue that standard model-predictive control algorithms face challenges performing station-keeping due to the complex non-linear relationship between control decisions and the target objective \cite{bellemare_autonomous_2020}.  As a result, the authors use Quantile Regression-based Deep Q-learning (Qr-DQN) \cite{dabney_distributional_2018}, parameterised using a neural network.  The agent's actions represent the intake from a pump or exhaust of air within the balloon envelope.  Furthermore, the authors model the thermal expansion of the balloon envelope and solar energy generated for a renewable resource for the pump.  Effectively learning the effect of the diurnal cycle.  Furthermore, a Gaussian Process (GP) was used to model the uncertainty between the wind forecast errors and the true wind speed.  Bellemare's work illustrated the large computation required to train a reasonable policy.   Xu {\it et al.} showed how beneficial a prioritised experience replay \cite{schaul_prioritized_2016} based on high-value samples can improve the training stability for station-keeping tasks \cite{xu_station-keeping_2022} whilst using a Deep Q-Network \cite{mnih_human-level_2015}.

Saunders {\it et al.} illustrated that reinforcement learning could also be used for latex balloons, a low-cost alternative to super-pressure balloons \cite{saunders_resource-constrained_2023}.    The use of sand ballast and venting helium enables latex balloons to reach much larger ascent rates compared to super-pressure balloons.  However, venting and ballasting come at a greater cost of resources, which is non-renewable.  The authors investigate resource-constrained station-keeping while incorporating feasible limits on the resources exhausted. 

Alternative balloon systems have been proposed to improve station-keeping performance.  Jiang {\it et al.} proposes a double-balloon system which uses a winch to control the altitude of an assistant balloon \cite{jiang_station-keeping_2020}.  This assistant balloon reaches desired altitudes using the winch to change the flight direction, using less energy than air ballasts.  Furthermore, Wynsberghe {\it et al.} proposes electro-hydrodynamic thrusters, which deliver power wirelessly from a ground-based transmitter \cite{van_wynsberghe_station-keeping_2016}.  However, the disadvantage of this approach is the requirement to be in the line of sight of the transmitter.

The uncertainty in modelling the wind and atmospheric effects is still an open research question \cite{sobester_high-altitude_2014, fields_-flight_2013}, which can lead to a large sim-to-reality gap.  Fields {\it et al.}  Attempts to overcome this gap using online wind data \cite{fields_-flight_2013} to more accurately predict the landing location.  Wind data is collected on the ascent phase, and used to correct the flight parameters on the descent.  Alternatively, S\'{o}bester {\it et al.} produces a balloon flight simulation model that considers an empirically derived stochastic drag model along with uncertainties in the wind profile and the balloon envelope \cite{sobester_high-altitude_2014}.  Then, Monte-Carlo ensembles to predict a trajectory along with the landing site with location error estimates.

Within the context of capacity optimisation, researchers have used similar techniques to optimise for cell tower coverage \cite{dreifuerst_optimizing_2021} and wind turbine placement \cite{asaah_optimal_2021}.  To our knowledge, we are the first to optimise the launch location of a latex high-altitude balloon in order to enhance the station-keeping performance.



\section{Background}\label{sec:bkgrnd}
Here we discuss the force balance of the latex balloon and the formulation for the reinforcement learning policy used for station-keeping.  Then we outline the background of the Gaussian Process and the Bayesian Optimisation employed to search for the optimal launch configuration.



\subsection{Equations of Motion}
For our study, we use the dynamic model of a latex balloon, where the equations of motion are explained below.  For further details, we direct the reader's attention to our previous work \cite{saunders_resource-constrained_2023}.
The forces acting on the balloon include the buoyancy $F_b$, weight $F_w$, and drag $F_d$ force, such that $\sum F = F_b - F_d - F_w$ \cite{farley_balloonascent_2005}.  The buoyancy force is a result of the displaced air caused by the balloon, $F_b = \rho_a V g$, where both the density of air $\rho$ and, as a consequence, the volume of the balloon envelope $V$ vary with altitude.  The drag force acts opposite to the relative motive of the balloon with respect to the wind $F_d = \frac{1}{2}\rho c_d A |\mathbf{v_r}|\mathbf{v_r}$ \cite{taylor_classical_2005}.  Given that $\mathbf{v_r}$ is the motion of the balloon $\mathbf{v_b}$ relative to the wind $\mathbf{v_w}$, such that $\mathbf{v_r} = \mathbf{v_b} - \mathbf{v_w}$. $A$ is the cross-sectional area, and $c_d$ is the drag coefficient.  Finally, the gravitational force is the combined weight of the inert mass $m_i$, including the payload and balloon envelope, helium mass $m_h$, and sand mass $m_s$.  Where the combined mass is $m = (m_p + m_h + m_s)$, and $F_w = mg$.  We represent the basis vector for which the weight and buoyancy forces act along as $\mathbf{e_z} = [0, 0, 1]^T$.
\begin{equation} \label{eq:N2L}
    m\mathbf{a} = \rho_a V g \mathbf{e_z}  - \frac{1}{2}\rho c_d A |\mathbf{v_r}|\mathbf{v_r} - (m_p + m_h + m_s)g\mathbf{e_z}
\end{equation}

Assuming helium acts as an ideal gas and the latex material acts perfectly elastic \cite{sushko_low_2017}, the balloon envelope volume can be calculated using $V=\frac{nRT}{P}$.  Where $P$ is the ambient pressure, $R$ is the universal gas constant, $n$ is the number of mols of helium, and $T$ is the ambient temperature.  The atmospheric variables are all modelled using the US Standard Atmosphere Model 1976.  Using atmospheric lapse rates $L$, the ambient temperature $T$, given a reference temperature $T_0$, can be calculated by $T = T_0 + (h \times L)$.  We assume the internal temperature is equivalent to the ambient temperature. Then, given the volume, we can calculate the drag area $A=\pi\left(\frac{3V}{4\pi}\right)^{\frac{2}{3}}$.

\subsection{Reinforcement Learning}
We formulate the task of station-keeping as a Markov Decision Process (MDP), characterized by the tuple $(\mathcal{S}, \mathcal{A}, P, R)$.  For each decision-step $t$, the agent in state $s_t\in\mathcal{S}$ transitions to a new state $s_{t+1}$ after taking an action $a_t\in\mathcal{A}$, guided by the probability distribution $P(s_{t+1}|s_t,a_t)$.  As a result of reaching $s_{t+1}$, the agent receives a reward $r_{t+1}$ according to the reward distribution $R(s_t, a_t)$.  The agent's objective during training is to maximise the expected future discounted cumulative reward $\mathbb{E}[G_t|s_t]$. Where $G_t=\sum^T_{k=t+1}\gamma^{k-t-1}R_k$ and $\gamma\in[0,1]$ is the discount factor.  

We make use of a Soft-Actor Critic (SAC) \cite{haarnoja_off-policy_2018} to control the balloon which is an off-policy deep RL algorithm based on the maximum entropy framework.  The policy aims to maximise both the expected reward and policy entropy, and simultaneously learns a policy $\pi_\theta$ and a Q-function $Q_{\phi}$, which are parameterised using a neural network.  Therefore, the Q-function parameters are optimised using the mean squared loss function $J_Q(\theta)=\mathbb{E}_{(s_t, a_t)\sim\mathcal{D}}\left[\frac{1}{2}\left(Q_\phi(s_t,a_t)-\hat{Q}(s_t,a_t)\right)^2\right]$, between the target $Q_\phi$ and predicted $\hat{Q}$ action-value.  Where samples are drawn from a replay buffer $\mathcal{B}$. Given that the predicted action-value equates to $\hat{Q}(s_t,a_t)= r(s_t,a_t)+\gamma(1-d)\mathbb{E}_{a\sim\pi}[ Q_\phi(s_{t+1},\pi_\theta(a_{t+1}|s_{t+1})) - \alpha\log\pi_\theta(a_{t+1}|s_{t+1})]$ and the entropy temperature $\alpha$ is adjusted by taking the gradient of $J_\alpha=\mathbb{E}[-\alpha\log\pi_\theta(a_t|s_t;a)-\alpha \bar{H}]$ \cite{haarnoja_soft_2019} with a desired minimum entropy $\bar{H}=0$.

The policy is optimised by minimising the simplified Kullback-Leibler divergence $J_\pi(\phi)=\mathbb{E}_{s_t\sim\mathcal{B},\epsilon_t\sim\mathcal{N}}[\log\pi_\phi(\tilde{a}(s_t, \epsilon_t)|s_t)-Q_\theta(s_t, \tilde{a}(s_t, \epsilon_t))]$.  Where samples from $\pi_\theta$ are computed using a squashed Gaussian policy using the reparameterisation trick $\tilde{a_t}_\theta(s, \epsilon)=\tanh(\mu_\theta(s)+\sigma_\theta(s)\odot\epsilon)$.  Where epsilon is sampled from a standard normal $\epsilon\sim\mathcal{N}(0, I)$.

\subsection{Spatial-Temporal Gaussian Process Modelling}
For this study, a GP is used to model the noisy spatial-temporal station-keeping performance of the balloon from trails collected from running the policy through a simulated episode.

A GP is a versatile Bayesian non-parametric method that can learn unknown non-linear functions by placing a prior distribution over the space of functions \cite{rasmussen_gaussian_2006}.  Noise observed target values $y$ are modelled as $y=f(\mathbf{x}) + \epsilon$ where $\epsilon \sim \mathcal{N}(0, \sigma^2\mathbf{I})$ is noise associated to each independent observation.  The GP is determined by a mean function $m(\textbf{x})$ and a positive semi-definite covariance function $k(\textbf{x},\textbf{x}')$ such that,

\begin{equation}
    f(\textbf{x})\sim \mathcal{GP}(0, K(\textbf{x}, \textbf{x}'))
\end{equation}

For this paper, we assume a zero mean function, $m(\textbf{x})=0$.  There are several covariance functions within the literature, each with different characteristics.  For our study, we make use of the $\nu=\frac{5}{2}$ Mat\'{e}rn kernel due to its robustness against non-smooth functions.
\begin{equation}
    K_{\text{Mat\'{e}rn}}(\mathbf{x}, \mathbf{x}') = \frac{2^{1 - \nu}}{\Gamma(\nu)} \left( \sqrt{2 \nu} d \right)^{\nu} K_\nu \left( \sqrt{2 \nu} d \right)
\end{equation}

Where $d$ is the distance between $x$ and $x'$ scaled by the lengthscale parameter $\theta$.  $\nu$ is the smoothness parameter, $\Gamma$ is the gamma function, $K_{\nu}$ is the modified Bessel function of the second kind.

Finding the optimal hyper-parameters $\theta_{\text{gp}}^*$ can be achieved by maximising the log marginal likelihood of the data.
\begin{equation}
\begin{aligned}
    \theta_{\text{gp}}^* &  = \max_\theta \log \left( - \frac{1}{2} \log |\textbf{K}| -\frac{1}{2} \mathbf{y}^T \textbf{K}^{-1} \mathbf{y} - \frac{n}{2} \log 2\pi \right)
\end{aligned}
\end{equation}

\subsection{Bayesian Optimisation}

To build the GP model, we use a Bayesian Optimisation (BO) method to estimate the next sample point.  To achieve this, BO utilises an acquisition function $h$, which guides the search over the GP.  The process involves finding the parameters $\mathbf{x}$ that maximises the acquisition function at each iteration, $\mathbf{x} = \argmax_{\mathbf{x}} h(\mathbf{x})$.

\begin{algorithm}
\caption{Generic Bayesian Optimization Algorithm}
\label{alg:BO}
\begin{algorithmic}[1]
\Require $f, h, \mathcal{D}$
\Ensure $\mathbf{x}^*, f(\mathbf{x}^*)$
\For{$j = 1, 2, 3, \ldots$}
    \State Find $\mathbf{x}_i = \argmax_\mathbf{x} h(\mathbf{x})$ \Comment{Max Acquisition function}
    \State $y_i \leftarrow f(\mathbf{x}_i)$ \Comment{Obtain sample from $f$}
    \State $\mathcal{D} \leftarrow \mathcal{D} \cup \{(\mathbf{x}_i, y_i)\}$ \Comment{Update training set}
    \State $\theta_{\text{gp}}^* = \max_\theta \log p(\mathbf{y}|\mathbf{x},\theta_{\text{gp}})$ \Comment{Update GP}
    \If{$y_j > \mu(\mathbf{x}^*)$}
        \State $\mathbf{x}^* \leftarrow \mathbf{x}_i$ \Comment{Update location of optimum}
    \EndIf
\EndFor
\end{algorithmic}
\end{algorithm}

For our study, we use the expected improvement acquisition function $h(\mathbf{x}) = \mathbb{E}[I(\mathbf{x})]$, given that improvement is $I(\mathbf{x})=\max(f(\mathbf{x})-f(\mathbf{x}^*), 0)$ where $f(\mathbf{x})$ denotes our current sampled value and $f(\mathbf{x}^*)$ our current best-sampled value.  The closed form can be expressed as the following \cite{jones_efficient_1998}.


\begin{equation}
\begin{aligned}
    h(x) & = (\mu-f(\mathbf{x}*))\Theta(\frac{\mu-f(\mathbf{x}*)}{\sigma})+\sigma\psi(\frac{\mu-f(\mathbf{x}*)}{\sigma})
\end{aligned}
\end{equation}

\looseness=-1 Where we emit the explorationtuning parameter, $\xi=0$.  The generic algorithm for BO follows Algorithm \ref{alg:BO}.

\section{Problem Statement}\label{sec:problem}

This paper aims to identify the launch configuration $\textbf{x}\in\{x_0, y_0, \Delta t\}$ such that the expectation of the undiscounted return for the given policy $\pi$ is maximised.  The launch configuration consists of the initial longitude $x_0$, latitude $y_0$, and time offset $\Delta t \in \mathbb{R}^+$ from a specific start date.  Here $\textbf{x}$ forms part of the initial state, $s_0\in\{..., ||(x_0, y_0)||,...\}$ and the wind vector sampled $v_w=W(x, y, z, t + \Delta t)$.  

For our study, we limit the spatial distance of the search space to $x_{\text{max}} = y_{\text{max}} = 400$\;km.  Furthermore, the time offset is limited to $\Delta t_{\text{max}} = 24$ hours, as longer offsets require further study into the variance of the wind forecast \cite{friedrich_comparison_2017}. The problem is formally defined as:

\begin{equation}
\begin{aligned}
    &\argmax_{x_0,
    \,y_0,\,\Delta t}\quad \mathbb{E}\left[ \sum^\infty_{t=0} R_{t+1} | S_t = s_0 \right] \\
    &\textrm{s.t.} \quad  |x_0| \leq x_{\text{max}},\quad|y_0| \leq y_{\text{max}} ,\quad \Delta t_0 \leq t_{\text{max}}\\
\end{aligned}
\end{equation}

\section{Method}\label{sec:method}

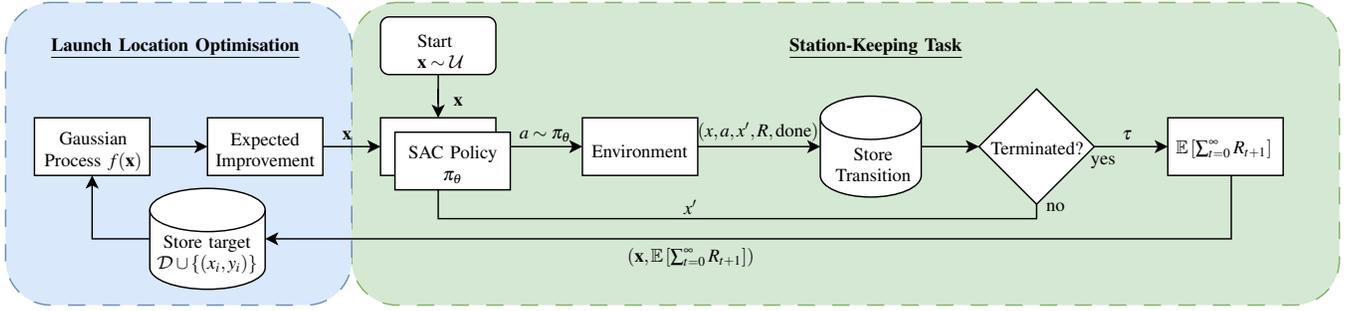
\begin{figure*}
\centering
\begin{overpic}[width=\linewidth]{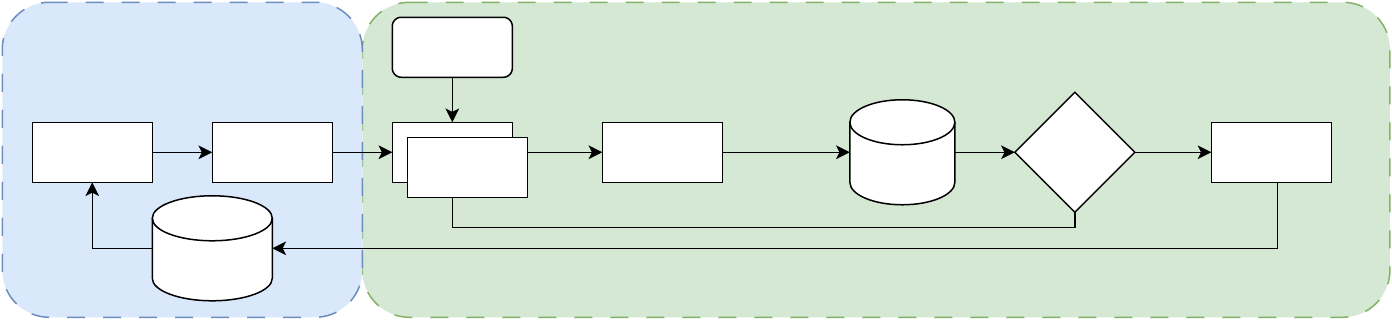}
    \put(34, 61){\makebox(0,0){\begin{minipage}{2cm}
    \scriptsize \centering Gaussian \\ Process $f(\textbf{x})$ \end{minipage}}}
    \put(99, 61){\makebox(0,0){\begin{minipage}{2cm}
    \scriptsize \centering Expected \\ Improvement \end{minipage}}}
    \put(170, 56){\makebox(0,0){\begin{minipage}{2cm}
    \scriptsize \centering SAC Policy \\ $\pi_\theta$ \end{minipage}}}
    \put(185, 99){\makebox(0,0){\begin{minipage}{2cm}
    \scriptsize Start \\ $\textbf{x} \sim \mathcal{U}$ \end{minipage}}}
    \put(251, 61){\makebox(0,0){\begin{minipage}{2cm}
    \scriptsize Environment \end{minipage}}}
    \put(329, 56){\makebox(0,0){\begin{minipage}{2cm}
    \scriptsize \centering Store \\ Transition \end{minipage}}}
    \put(402, 62){\makebox(0,0){\begin{minipage}{2cm}
    \scriptsize Terminated? \end{minipage}}}
    \put(472, 62){\makebox(0,0){\begin{minipage}{2cm}
    \scriptsize $\mathbb{E}\left[ \sum^\infty_{t=0} R_{t+1} \right]$ \end{minipage}}}
    \put(77, 22){\makebox(0,0){\begin{minipage}{2cm}
    \scriptsize \centering Store target \\ $\mathcal{D} \cup \{(x_i, y_i)\}$ \end{minipage}}}

    \put(260, 40){\makebox(0,0){\scriptsize $x'$}}
    \put(260, 21){\makebox(0,0){\scriptsize $(\textbf{x}, \mathbb{E}\left[ \sum^\infty_{t=0} R_{t+1} \right])$}}
    \put(416, 57){\makebox(0,0){\scriptsize yes}}
    \put(398, 40){\makebox(0,0){\scriptsize no}}
    \put(425, 67){\makebox(0,0){\scriptsize $\tau$}}
    \put(205, 67){\makebox(0,0){\scriptsize $a \sim \pi_\theta$}}
    \put(285.5, 68.5){\makebox(0,0){\scriptsize $(x, a, x', R, \text{done})$}}
    \put(130, 67){\makebox(0,0){\scriptsize $\mathbf{x}$}}
    \put(172, 80){\makebox(0,0){\scriptsize $\mathbf{x}$}}

    \put(330, 100){\makebox(0,0){\scriptsize \underline{\textbf{Station-Keeping Task}}}}

    \put(65, 100){\makebox(0,0){\scriptsize \underline{\textbf{Launch Location Optimisation}}}}

\end{overpic}
\vspace*{-6mm}
\caption{\small Block diagram of our method to identify the optimal launch configuration.  Initially, we randomly sample a launch configuration which forms part of the initial state $x$ of the RL policy.  After the episode has terminated, the expected undiscounted return is calculated and stored along with the launch configuration $\mathbf{x}$.  Then, a GP models the performance of the policy by optimising the log marginal likelihood.  The next launch configuration is chosen by maximising the expected improvement.}
\label{fig:BlockDiagramBO}
\end{figure*}


In this section, we formulate the Soft Actor-Critic policy used for station-keeping, detailing the state and action space.  We then describe the data utilised for training and testing.  Then, we propose a reward function designed to improve the generalisation of unseen data and reduce reward hacking.  Finally, we outline how Bayesian optimisation maximises the expected improvement to identify the optimal launch configuration. We outline the components of our method in Figure \ref{fig:BlockDiagramBO}.

\subsection{Soft Actor-Critic Controller}
The Soft Actor-Critic policy, which controls the balloon, is parameterised by a fully connected neural network as illustrated in our previous work \cite{saunders_resource-constrained_2023}.  Both actor and critic networks have two hidden layers of 256 neurons each.  The actor network is parameterised by a Gaussian policy, where actions are represented as the hyperbolic tangent Tanh applied to z values sampled from the mean and covariance given by the network.  Meanwhile, the critic is modelled as a soft Q-function.

The MDP state space ($\mathcal{S}$) consists of a collection of wind and ambient features.  Wind features, consisting of magnitude $|v|$ and bearing error $\theta$, are sampled at 25 equally-spaced points, between the vertical pressure limits $[5000, 14000]$\, Pa of the forecast.  The ambient features consist of onboard measurements, which include the altitude $h_t$, ascent rate $\dot{h}_t$, balloon envelope drag area $A$ and volume $V$, helium mols $n_h$, and total system $m_T$ and sand $m_s$ mass.  Furthermore, the wind velocity $|v_h|$ and bearing error at the current altitude $\theta_h$, and distance $d$ and heading to the target $[\sin(\theta_x), \cos(\theta_x)]$ also form part of the ambient features.  Where $d=||(x,y)||$ is the distance at the current decision step to the target, which has a radius $r$.  The past three altitudes $[h_{t-1}, h_{t-2}, h_{t-3}]$, ascent rates, $[\dot{h}_{t-1}, \dot{h}_{t-2}, \dot{h}_{t-3}]$, and float actions $[a_{2, t-1}, a_{2, t-2}, a_{2, t-3}]$ are included to incorporate agent memory.  Both wind and ambient features are concatenated into a single vector of length $77$.

The MDP action space ($\mathcal{A}$) consists of three actions $a\in[a_0, a_1, a_2]$.  Consisting of desired altitude $a_0\in[14, 21]$\; Km,  time-factor $a_1\in[1, 5]$, and finally, if to float $a_2\in[-1, 1]$.  The desired ascent rate is calculated $\dot{h}_d$:
\begin{equation}
    \dot{h}_d=\mathds{1}_{a_2\in[-1, 0]}\left(\frac{a_0 - h_t}{a_1 \times T}\right)
\end{equation}

Where $\mathds{1}$ represents the indicator function. Then, given the desired ascent rate, the desired sand ballast can be calculated if the desired ascent rate is larger than the current ascent rate, $\dot{h}_d > \dot{h}_t$.
\begin{equation}
m_{\text{s,calc}} = \rho V - \frac{1}{2g} C_d A \left|\dot{h}_d\right|\dot{h}_d - m_p - m_h 
\end{equation}
Or, the desired vented helium can be calculated if the desired ascent rate is less than the current ascent rate, $\dot{h}_d < \dot{h}_t$.  Where we solve $n_{\text{calc}}$ in,
\begin{align}
\begin{split}
    \rho g \left(\frac{RT}{P} - M\right)n_{\text{calc}}
    - \frac{1}{2}\rho\left|\dot{h}_d\right|\dot{h}_d&C_d\pi\left(\frac{3RT}{4\pi P}\right)^{\frac{2}{3}}n_{\text{calc}}^\frac{2}{3} \\
    & -\left(m_p + m_s\right)g=0
\end{split}
\end{align}

\subsection{Wind Data}

We make use of the ECMWF's ERA5 global reanalysis dataset \cite{hersbach_era5_2020}.  The wind vectors are located over a grid of points in a parameter space of $\mathcal{X} \times \mathcal{Y} \times \mathcal{P} \times  \mathcal{T}\times  \mathcal{V}$.  The longitude $\mathcal{X}$ and and latitudinal $\mathcal{Y}$ positions are sampled with a resolution of $0.4^{\circ}$ at pressure points $\mathcal{P}$ ranging from $2000$\,Pa to $17500$\,Pa.  Finally, the wind fields have a time separation of $6$ hours and are collected between $1$st November $2022$ and $28$th February $2023$ at longitude $-113^{\circ}$ latitude $1^{\circ}$.  The data is split into train and test, such that the training set consists of dates between $1$st November $2022$ to $31$st January $2023$ and the test set contains dates between $1$st February $2023$ to $28$th February $2023$.  Furthermore, simplex noise \cite{perlin_image_1985} is used to augment the wind forecast to emulate forecasting errors \cite{coy_global_2019}. 

\subsection{Reward Function Evaluation}

The effectiveness of the policy generalising to unseen data can be influenced by the reward function.  A trade-off exists between the risk of reward hacking with ambiguous reward functions and the potential over-constraint policy behaviour with overly defined functions.  Contextually, the agent could overfit to specific idiosyncrasies of the environment, such as weather patterns, and, if not entirely constrained, could also disregard theoretical venter or ballast limits \cite{saunders_resource-constrained_2023}.

Previous studies employed reward functions which provided no distinction of state values within the region.  We find, for latex balloons, this approach lead the agent to navigate close to the circumference of the region, frequently causing it to approach the region by flying around the perimeter.  Therefore, we augment the previously used Step reward function $R_{\text{Step}}$ \cite{bellemare_autonomous_2020, saunders_resource-constrained_2023}, by incorporating a Tanh function $R_{\text{Tanh}}$.  This augmentation provides further incentives to fly closer to the centre radius.

The Step function, used first by Bellemare {\it et al.} illustrates a cliff edge, with a cliff edge constant of $c=0.4$ to provide an immediate distinction between inside and outside the region.  Additionally, a decay function is used to give higher rewards to states closer to the target. Given that $d - \rho$ is the distance away from the target and $\tau$ is the half-life.

\begin{equation}
\label{eq:reward_function}
R_{\text{Step}}=
\begin{cases}
    1.0, & \text{if }d < r \\
    c\times2^{-(d-\rho)/\tau)}, & \text{otherwise}
\end{cases}
\end{equation}

We make a slight adjustment to the Step reward function to encourage the agent to fly closer to the centre. As mentioned previously, states on the border of the region are inadvertently risky.  Therefore, Tanh is used instead of the singular $+1.0$.  We adjust the Tanh function's scaling to ensure the maximum reward at the centre is $+1.0$, which then progressively decreases to $0.72$ at the boundary, maintaining a clear distinction between the two areas.

\begin{equation}
    R_{\text{Tanh}}=
\begin{cases}
    -(\tanh((d/20) - 3) - 1) / 2, & \text{if }d < r \\
    c\times2^{-(d-\rho)/\tau)}, & \text{otherwise}
\end{cases}
\end{equation}

Two separate policies are trained using the two reward functions $R_{\text{Step}}$ and $R_{\text{Tanh}}$.  With both policies trained using the same seeded environment and the same MDP formulation.  Launch locations during training are randomly sampled uniformly across a 400\;km radius $r \sim \mathcal{U}(-400, 400)$ and $\theta \sim \mathcal{U}(0, 2\pi)$.  

To evaluate how well the policy generalises to unseen data, we evaluate both policies on the test dataset.  We calculate the average steps within region over all trajectories as $\frac{1}{N\times T}\sum^N_{n=0}\sum^T_{t=0}\mathds{1}_{d_t<r}$.  Where $N$ represents the total number of trajectories, and $T$ represents the total number of decision steps. We also calculate the ratio of trajectories passing through region $\frac{1}{N}\sum_n^N \mathds{1}_{d_t < r}$.  The average time within region illustrates the average performance of the policy, whereas the number of trajectories passing through the region indicates the spread of viable launch locations.


\subsection{Launch Location Optimisation}



BO is used to build a GP model representing the spatial-temporal performance of each trained policy given the two reward functions: $R_{\text{Step}}$ and $R_{\text{Tanh}}$.

An initial sample location, as previously done, is randomly sampled from a uniform distribution $(x_0, y_0) \sim \mathcal{U}(-400, 400)$, $\Delta t \sim \mathcal{U}(0, 24)$.  For each episode, actions are selected by taking the mean of the GP $\mu_\theta(s)$ where each transition $(s, a, s', r)$ is saved until a terminal condition is reached.  The expected undiscounted return received throughout the trajectory is calculated, $\mathbb{E}[\sum_{t=0}^T R_{t+1}]$, which acts as the target, along with the initial launch parameter $\mathbf{x}\in(x_0, y_0, \Delta t)$.  Both the target and sample are appended to the GP training set $\mathcal{D} \leftarrow \mathcal{D} \cup \{(\mathbf{x}, \mathbb{E}[\sum_{t=0}^T R_{t+1}])\}$.  The GP is then updated with this new sample point, by maximising the log marginal likelihood $\max_\theta \log p(\mathbf{y}|\mathbf{x},\theta_{\text{gp}})$.  The next sample point is chosen which maximises the expected improvement $\mathbb{E}[I(\mathbf{x})]$.  This next sample point is used as the initial state of the balloon.

This procedure is performed for every day within the test set, $1$st to $28$th February, to evaluate the best location and time-offset to launch the balloon.  To evaluate the performance of BO, we compare the average converged score and convergence time for all days within the test set against Particle Swarm Optimisation (PSO) and by uniformly sampling over the space.   Given that the average convergence score is defined as $\frac{1}{N}\sum_{n=0}^N \textbf{x*}$ where in this instance $n$ represents the date in the test set.  Furthermore, the average converged index is defined as $\frac{1}{N}\sum_{n=0}^N \argmax_\textbf{x*}$.  We set a maximum number of steps to 1500 for each day within the test set.  We compare BO against PSO and uniform sampling.  To give PSO a fair chance, we linearly decay the cognitive $c_1$, social $c_2$, and inertia $w$ such that $w = 0.4\frac{n_i-N_i}{N_i^2}+0.4$, $c_1 = -3\frac{n_i}{N_i}+3.5$, $c_2 = 3\frac{n_i}{N_i} + 0.5$ \cite{shi_modified_1998}.  Where $n_i$ and $N_i$ are the current index and maximum index respectively.  The initial values for both the cognitive and social parameters are set at one, $c_1 = c_2 = 1$, and the inertial weight is $w=0.8$.


\section{Results}\label{sec:results}

We present our results that indicate that the Tanh reward function generalises to the unseen test data better than the Step function.  Furthermore, we illustrate that BO is able to locate an optimal launch configuration in less time compared to PSO and uniform sampling.

\subsection{Reward Function Evaluation}

\begin{figure}[tbp]
\centering
\includegraphics[width=\linewidth]{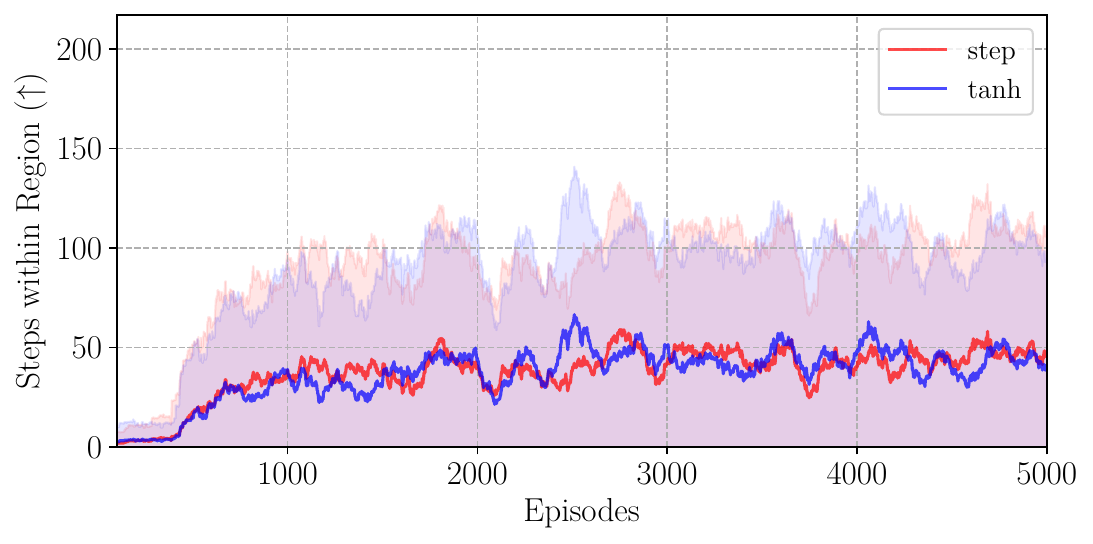}
\vspace*{-7mm}
\caption{\small Training curve of the two policies, indicating performance converging to approximately 45 steps within region.  Illustrating similar performance to the training data.}
\label{fig:training_of_different_rewards}
\end{figure}

Both policies during training converge, on average, to the same time within region, as illustrated in Figure \ref{fig:training_of_different_rewards}.  Previous studies have pointed out the diminishing returns after longer training times, with some studies training a policy for weeks \cite{bellemare_autonomous_2020}.  Hence, we stop training after 5000 episodes, given that both policies have converged.

\begin{figure}[bp]
\centering
\includegraphics[width=\linewidth]{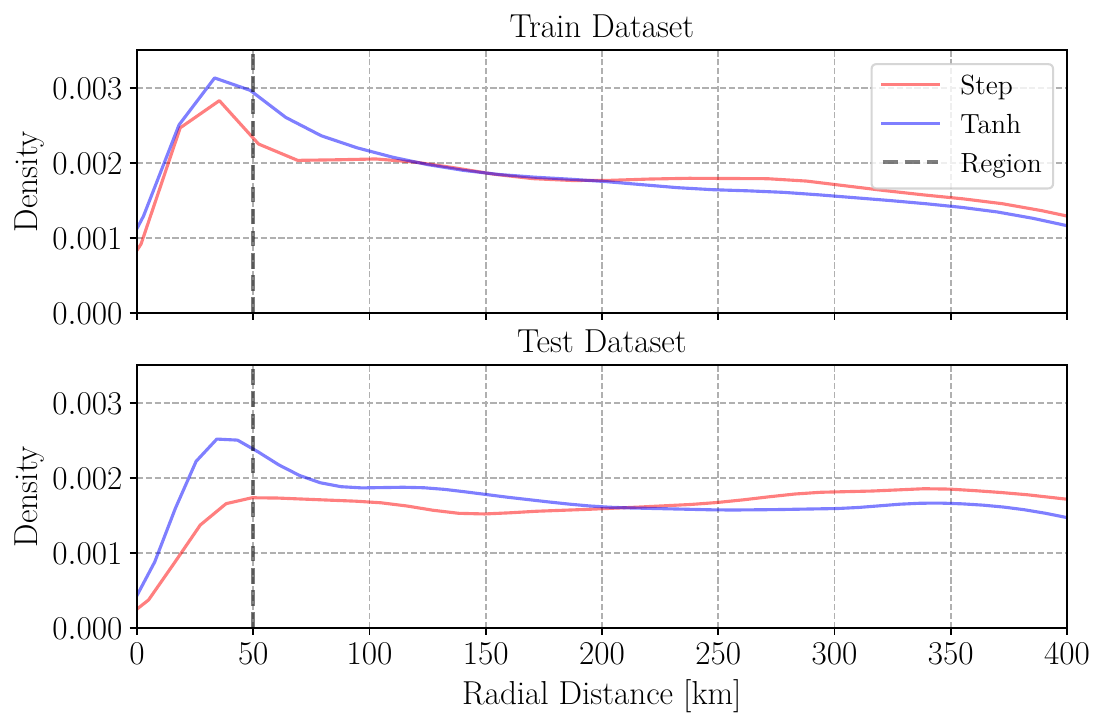}
\vspace*{-7mm}
\caption{\small Kernel density estimate visualising the distribution of positions for both policies between 0 and 400\;k.  Indicating the worse generalisation for the Step function to unseen data.}
\label{fig:kde_train_eval}
\end{figure}

The average time within the 50\;km radius and the ratio of trajectories reaching the region in unseen test data is illustrated in Table \ref{tab:reward_comparison_table}.  The policy trained using a Tanh reward function achieved a higher average score on unseen data.  Therefore, it was found that this policy was able to generalise better to unseen data.  Furthermore, the Tanh reward function illustrated a higher proportion of trajectories reaching the region.  To illustrate this further, we plot a kernel density estimate which visualises the distribution of locations reached by the balloon in Figure \ref{fig:kde_train_eval}.

\begin{table}[bp]
    \centering
    \begin{tabular}{|l|c|c|} \hline 
         & Average Steps & Ratio of Trajectories\\
 Reward Function& Within Region (↑)& Reaching the Region (↑)\\
 \hline \hline 
         Step &  14.11&0.29\\ 
         Tanh &  \textbf{23.20} & \textbf{0.49}\\ \hline
    \end{tabular}
    \caption{\small Test data results of average time within the 50\;km region and the ratio of trajectories reaching the region for both policies.}
    \label{tab:reward_comparison_table}
\end{table}

\begin{figure}[tbp]
\centering
\includegraphics[width=\linewidth]{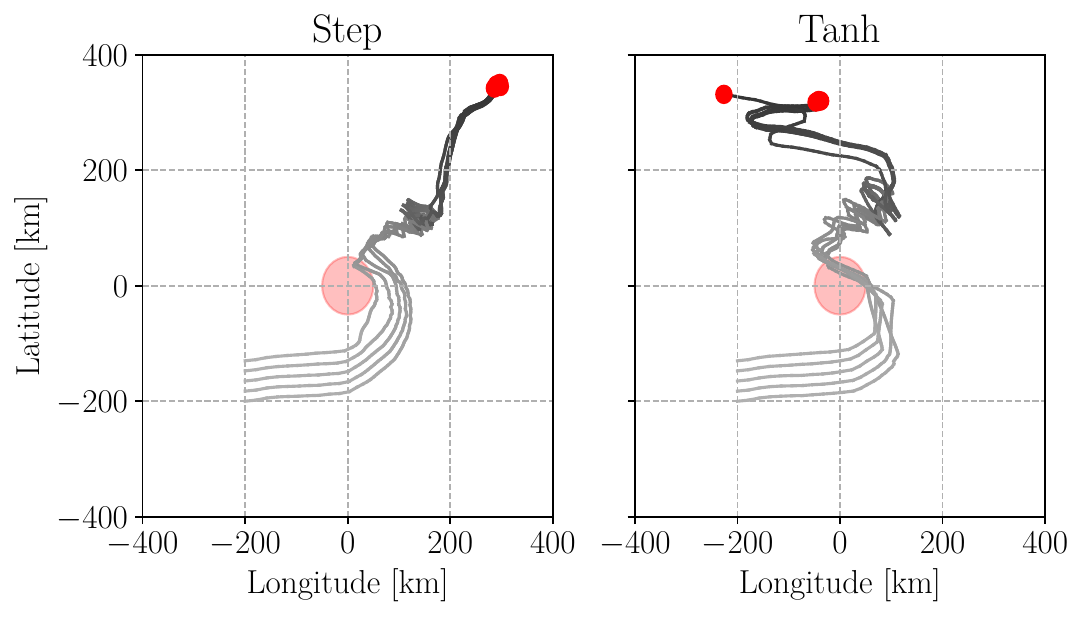}
\vspace*{-6mm}
\caption{\small Projected view of both policies trajectories given the same environment state with wind fields chosen from the test dataset.  Where initial positions are chosen at the same longitude initial position with varying latitudes.  The trajectories indicate the Tanh reward function incentivises the agent to fly closer to the target region.  The policy trained on the Step function attempts to traverse the circumference of the region which leads to fewer steps within the region.}
\label{fig:main_day1_-200000x_-200000y}
\end{figure}

Expectedly, the plot illustrates a decrease in the
distribution for both policies on the unseen test dataset relative to the training dataset. Notably, the policy trained with the Step function shows a greater reduction in density.  To clarify the observed behaviour, we plot the trajectories of both policies in Figure \ref{fig:main_day1_-200000x_-200000y}.

The policy trained with the Step function inadvertently maximises its reward by exploiting the environment.  More steps within the region, and hence more reward, were achieved by circumnavigating the region's circumference rather than traversing the diameter.  Alternatively, the Tanh reward function incentivised flying closer to the target location, thus creating a higher separation between the bounds of the region and the balloon.  This leads to a higher proportion of trajectories reaching the region and an, on average, higher time within region score.

\begin{figure*}[t]
\centering
\includegraphics[width=\linewidth]{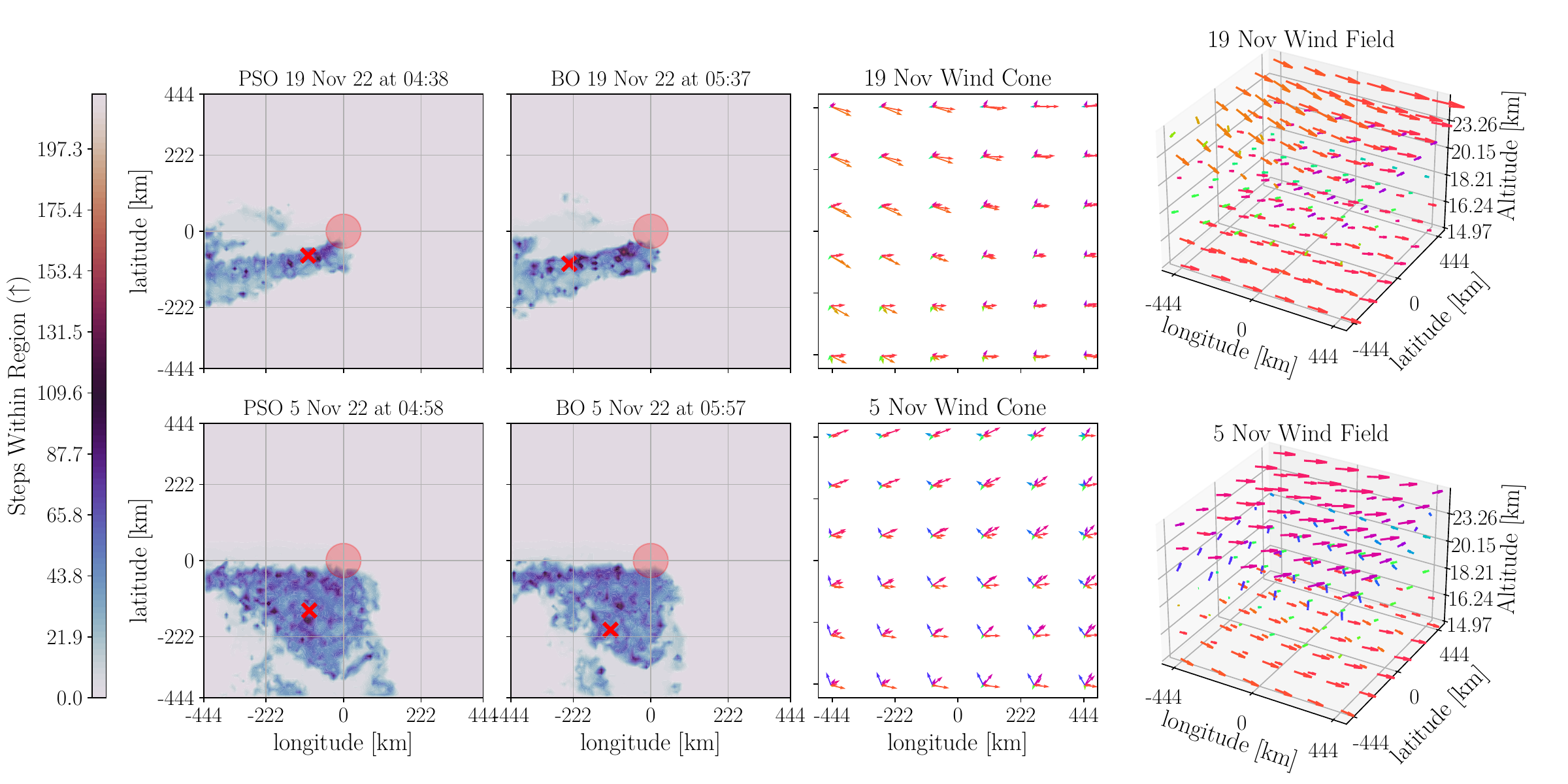}
\vspace*{-6mm}
\caption{\small Illustrating optimal positions, indicated by the red cross, across two different wind field dates. It can be observed that both PSO and BO identify subtly different launch configurations, spatially and temporally. Furthermore, it is obvious that the wind field has a significant effect on the performance, as shown by the wind cone.  Where the wind cone illustrates the diversity of the wind vectors at that position.}
\label{fig:optim_algo_two_diff_days}
\end{figure*}

\subsection{Launch Location Optimisation}

\begin{figure}[htbp]
\centering
\vspace*{-4mm}
\includegraphics[width=\linewidth]{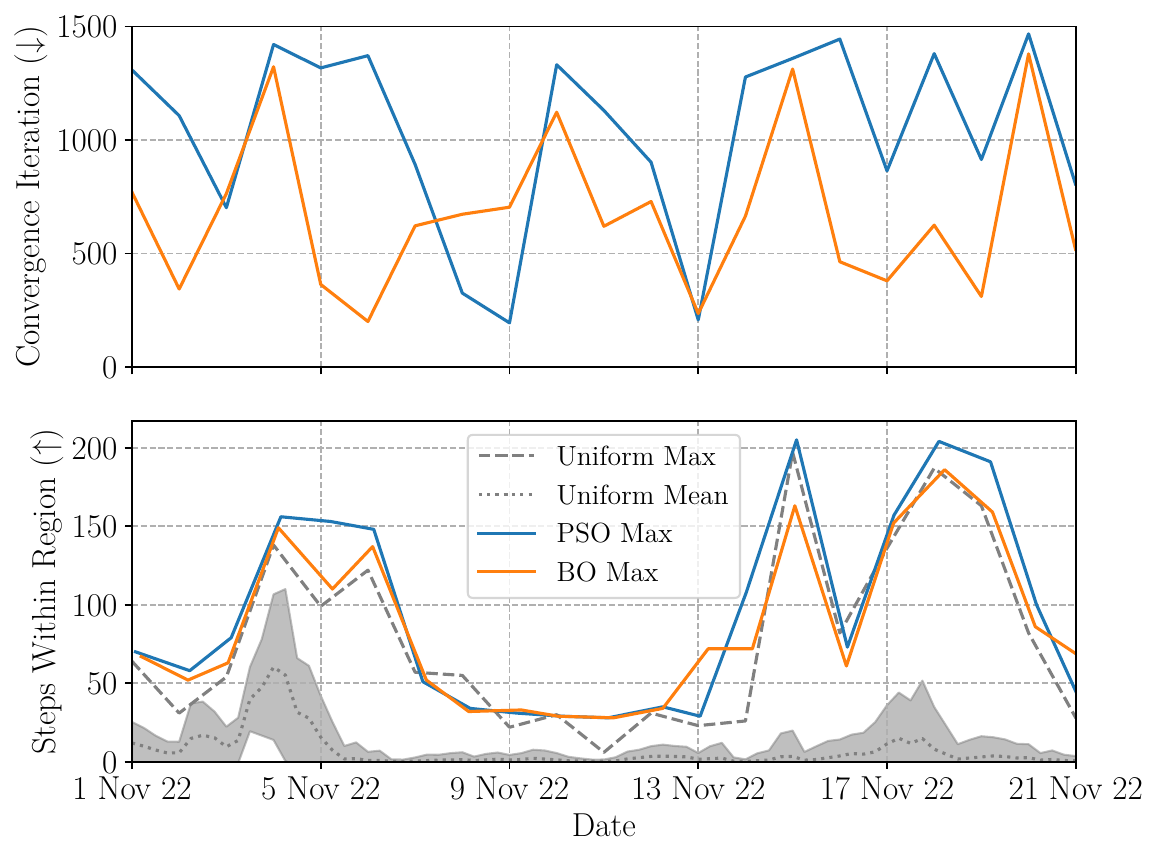}
\vspace*{-7mm}
\caption{\small Illustrates the converged index and station-keeping performance within the train dataset.  Showing that BO converges to the optimal performance in fewer steps than PSO.  All optimisation methods identify, on average, similar optimal performances.  The range of values achievable is indicated by the standard deviation obtained by uniformly sampling.}
\label{fig:tw50_over_all_dates}
\end{figure}

Average time within region, as illustrated by Figure \ref{fig:tw50_over_all_dates}, is not a good indicator for the maximum performance of the latex balloon.  The effect of seasons and diurnal cycles on performance has already been illustrated in previous literature \cite{bellemare_autonomous_2020}.  There exists a clear gap in the analysis of spatial performance for station-keeping.  Take the dates in Figure \ref{fig:tw50_over_all_dates}, the most optimal time to launch given the average time within region would be between $2^{\text{nd}}-6^{\text{th}}$ November.  However, given the maximum performance, the best time would instead be between the $17^{\text{th}}-20^{\text{th}}$ November.  Clearly, the wind field pattern has a significant effect on this performance as illustrated in Figure \ref{fig:optim_algo_two_diff_days}.

The convergence time and the maximum converged index for BO, PSO, and uniformly sampled for the test dataset is shown in Table \ref{tab:Optim_Comparison}.  Furthermore, the table is split into evaluation metrics for both reward functions.  All methods consistently converge to a similar maximum on average. However, BO converges in half the iterations required by other methods. Additionally, the Tanh method consistently attains a higher average TW50 score.

\begin{table}[htbp]
    \centering
    \begin{tabular}{|c|c|c|}
        \hline
         Method&  Average Converged & Average 
         Converge\\ 
          & Max (↑) & Index (↓) \\
         \hline \hline
         BO + Step& 68.42& \textbf{573.36}\\
         PSO + Step& 69.97&669.87\\
         Uniform + Step& 67.78&805.11\\
         \hline \hline
         BO + Tanh& 86.82&\textbf{459.32}\\
         PSO + Tanh& 89.71&976.03\\
         Uniform + Tanh& 83.68&787.61\\
         \hline
    \end{tabular}
    \caption{\small Evaluation metrics for all optimisation methods over the test dataset.  BO reaches the maximum converged score in fewer steps compared to PSO and Uniform.  Furthermore, Tanh reaches a higher average maximum time within region.}
    \label{tab:Optim_Comparison}
\end{table}

By analysing the optimal positions across two different wind field dates, as depicted in Figure \ref{fig:optim_algo_two_diff_days}, we observe that both PSO and BO identify subtly different launch configurations in both spatial (longitude and latitude) and temporal (launch offset) aspects.
The contour plots illustrate the dependence on the wind field and can be easily visualised with the wind cone.   Where the wind cone is obtained by projecting onto the latitude-longitude plane.  The wind cone is a good indicator of the diversity of wind vectors in the altitude layers.  Large wind cones indicate diverse wind vectors, creating better conditions for station-keeping.
Contrary to expectations, the chosen launch location is often outside the target region.

\section{Conclusion}\label{sec:conclusions}

High-altitude balloons have shown promise for their potential in atmospheric analysis and communication relay tasks, particularly through station-keeping. Yet, the efficacy of station-keeping is significantly influenced by atmospheric conditions, and determining the optimal launch location remains an open research question.  This challenge is particularly pronounced for latex balloons, which have a shorter flight time compared to their super-pressure counterparts.
To our knowledge, we are the first to address the problem of identifying the optimal launch location.  This problem involves identifying the most optimal time and location to launch the balloon in order to optimise the total steps within the target region. 
This problem is multifaceted and relies on underacted balloon dynamics driven by atmospheric conditions.
As done in previous works, we use a Soft Actor-Critic policy to control the balloon for station-keeping, and illustrate the problem, of misaligned behaviour at distances much further away from the target, which can limit the search space.  
This misaligned behaviour leads to less maximum reward and poorer generalisation of unseen wind fields.  
We propose optimising the agent by incentivising the agent to fly closer to the center, achieved using a Tanh function, and through experiments we find this new reward function achieves more steps within 50\;km in unseen data.
We illustrate how Bayesian Optimisation can identify optimal launch locations in fewer steps compared to other optimisation methods.  Furthermore, we show that the proposed reward function finds launch configurations which achieve, on average, higher steps within region.  Future research will focus on examining the impact of spatial constraints to emulate restricted regions and applying our methodology to a variety of other balloon models, such as super-pressure balloons.





\bibliographystyle{ieeetr}
\bibliography{references.bib}

\end{document}